\documentclass{article}

\usepackage{arxiv}
\usepackage[utf8]{inputenc}
\usepackage[T1]{fontenc}
\usepackage[hidelinks,hypertexnames=false]{hyperref}
\usepackage{url}
\usepackage{booktabs}
\usepackage{amsmath,amssymb,amsfonts}
\usepackage{graphicx}
\usepackage{natbib}
\usepackage{microtype}
\usepackage{tabularx}
\usepackage{array}
\usepackage{longtable}
\usepackage{threeparttable}
\usepackage{float}

\graphicspath{{figures/}}

\newcommand{\qninety}{q_{0.9}}
\newcommand{\qfifty}{q_{0.5}}

\title{Joint Upper-Bound Coverage and Route-Choice Utility:\\
An Empirical Evaluation on Two Urban Proxy Tasks}

\author{Jianing Long\textsuperscript{*}\\
School of Transportation Engineering\\
East China Jiaotong University\\
\texttt{ljnvs@foxmail.com}
\And
Xiaobin Li\\
School of Transportation Engineering\\
East China Jiaotong University
\AND
Wuming Lei\\
School of Transportation Engineering\\
East China Jiaotong University
\And
Weiguang Wang\\
School of Transportation Engineering\\
East China Jiaotong University}

\date{October 2, 2026}
\renewcommand{\shorttitle}{Upper-Bound Coverage and Route-Choice Utility}
\hypersetup{
  pdftitle={Joint Upper-Bound Coverage and Route-Choice Utility: An Empirical Evaluation on Two Urban Proxy Tasks},
  pdfauthor={Jianing Long; Xiaobin Li; Wuming Lei; Weiguang Wang},
  pdfsubject={Traffic forecasting, joint calibration, and route-choice evaluation},
  pdfkeywords={traffic forecasting; uncertainty calibration; joint coverage; reliable routing; decision evaluation}
}
\begin{document}
\maketitle

\begin{abstract}
Whether more accurate traffic forecasts or higher uncertainty coverage improve route decisions is unclear. We evaluate this question with a frozen protocol that separates speed error, joint candidate-path upper-bound coverage, route selection, and realized loss. Using processed road-speed data from Beijing and Chengdu, we construct offline proxy tasks with 150 origin--destination pairs, three candidate paths, and 14 test days per city. We compare raw 90th-percentile path-time bounds (C0) with jointly calibrated upper bounds (C2) under minimum-bound route choice. Joint coverage rises from 83.19\% to 92.26\% in Beijing M1, from 75.14\% to 88.33\% in Chengdu M1, and from 74.01\% to 90.64\% in Chengdu M2. Yet C2 increases lateness by 0.1633, 0.7848, and 0.9200 percentage points, respectively, and mean travel time by 0.588, 3.082, and 4.418 seconds. In a separate Chengdu predictor comparison, a 14.91\% reduction in speed mean absolute error accompanies a 1.4571-percentage-point reduction in lateness under C0. Joint coverage is therefore not a surrogate for downstream route utility in these frozen tasks; the offline results do not establish online or causal benefits.
\end{abstract}


\section{Introduction}\label{sec:intro}

\subsection{Motivation}
Traffic-speed forecasting has expanded from point estimates to predictive distributions, quantiles, and conformal methods. In route choice, a forecast is an input to a decision. Link-level uncertainty must be propagated to candidate-path travel times before a system chooses among paths using a reliability or risk objective. Lower average prediction error need not improve path ranking; likewise, wider upper bounds may cover realized travel times more often while changing the ranking of candidate paths and worsening travel time or lateness.

This distinction is particularly relevant in a traffic network. Candidate paths can share links, link-speed disturbances are correlated, and travel times vary with departure time. Marginal coverage for a link or individual path is not joint coverage for the set from which a path is selected. Moreover, the optimization interfaces $\arg\min U$, $\arg\min \mathbb{E}[T]$, and $\arg\min \Pr(T>B)$ express different decision objectives. An empirical comparison must therefore identify the calibration target, the rule that consumes the calibrated forecast, and the loss used to evaluate the selected route.

Prior traffic work provides rich forecasting and uncertainty methods, including deep sequence prediction with Q-Traffic data \citep{liao2018qtraffic}, unified views of traffic forecast uncertainty \citep{qian2024unified,wu2024adaptive}, and recent probabilistic traffic models \citep{xiong2026stochasticity}. Conformal and distribution-calibration research addresses correspondence between predicted uncertainty and observations \citep{romano2019conformalized,stankeviciute2021conformal,gibbs2021adaptive,xu2023sequential}. Decision-aware uncertainty methods ask how uncertainty affects actions \citep{cortesgomez2025utility,kiyani2025decision}. Reliable-routing studies address mean--risk trade-offs, dependent random link times, and adaptive paths \citep{samaranayake2012tractable,nikolova2014meanrisk,huang2018trajectory,filipovska2021reliable}. A common evaluation of coverage, route choice, and realized loss on fixed candidate sets remains useful because improvements at those layers need not agree.

\subsection{Questions and contributions}
We ask four questions. First, does a comparison between forecasting configurations with lower point error also show improved decisions under a fixed raw-upper-bound interface? Second, does joint candidate-set calibration move joint coverage toward a nominal 90\% target, and how much does it expand bounds? Third, after the calibrated bounds pass through a route-choice rule, how do lateness and travel time change? Fourth, what do route switches and their proximity to decision boundaries reveal about the observed changes?

The paper makes three contributions. It specifies a layered evaluation chain that separates forecast error, path-time scenarios, joint coverage, decision rules, and realized loss. It evaluates that chain on two public-data proxy tasks with fixed candidate sets and replayed truth, reporting absolute coverage, deviation from the 90\% target, bound expansion, simple baselines, oracle headroom, and decision effects. Finally, it separates prespecified formal comparisons from a post-result full-support branch and decomposes harmful and beneficial route switches. The contribution is an empirical evaluation and decision diagnostic, with explicit limits arising from processed data, short test periods, proxy paths, and the absence of causal identification.

\section{Related Work and Positioning}\label{sec:related}

\subsection{Traffic forecasting and uncertainty assessment}
Point forecasts are commonly assessed with mean absolute error (MAE), root mean squared error, or relative error. Distributional forecasts add quantile loss, interval coverage, reliability curves, and proper scoring rules. For dependent traffic time series and changing distributions, temporal and adaptive conformal approaches suggest more cautious evaluation than an independent-and-identically-distributed idealization \citep{stankeviciute2021conformal,gibbs2021adaptive,xu2023sequential}. These tools assess how uncertainty statements align with observations; they do not by themselves determine the value of a specific route-choice rule.

We do not propose a new forecasting network or claim full-distribution calibration. The object evaluated here is a finite candidate-set interface. Scenario travel times produce raw path quantiles and scales; a joint calibration adjusts candidate-path upper scores; the scores then enter an explicit route selector. This structure makes it possible to locate a mismatch between probabilistic and decision metrics.

\subsection{Prediction and decisions in transport networks}
Predict-then-optimize and decision-focused learning evaluate forecasts through downstream objectives \citep{donti2017task,wilder2019melding,elmachtoub2022smart}. Reliable routing under stochastic, correlated, and time-varying travel times shows why a route minimizing expected time need not minimize a risk measure \citep{samaranayake2012tractable,nikolova2014meanrisk,huang2018trajectory,filipovska2021estimation}. Rather than train an end-to-end decision model, we hold the route task fixed and evaluate the links between prediction, calibration, selection, and replay. The design supports diagnosis within a frozen task; it cannot substitute for online deployment, behavioral response, or a policy experiment.

\section{Evaluation Framework}\label{sec:framework}

\subsection{From link speeds to candidate-path distributions}
For origin--destination pair $o$, let $\mathcal{P}_o=\{p_1,p_2,p_3\}$ be the fixed candidate set. At departure time $t$ and residual scenario $b$, path travel time is
\begin{equation}\label{eq:path-time}
T_{o,p}^{(b)}(t)=\sum_{e\in p}\frac{\ell_e}{v_e^{(b)}(t)},
\end{equation}
where $\ell_e$ is link length and $v_e^{(b)}(t)$ combines predicted speed with a correlated residual scenario. The scenario set yields a path median $\qfifty(T_{o,p})$, raw 90th percentile $\qninety(T_{o,p})$, and task-specific scale $s_{o,p}$. The scale describes dispersion of scenario path times within this task; it is not a separately validated physical risk parameter.

\begin{figure}[t]
  \centering
  \includegraphics[width=\linewidth]{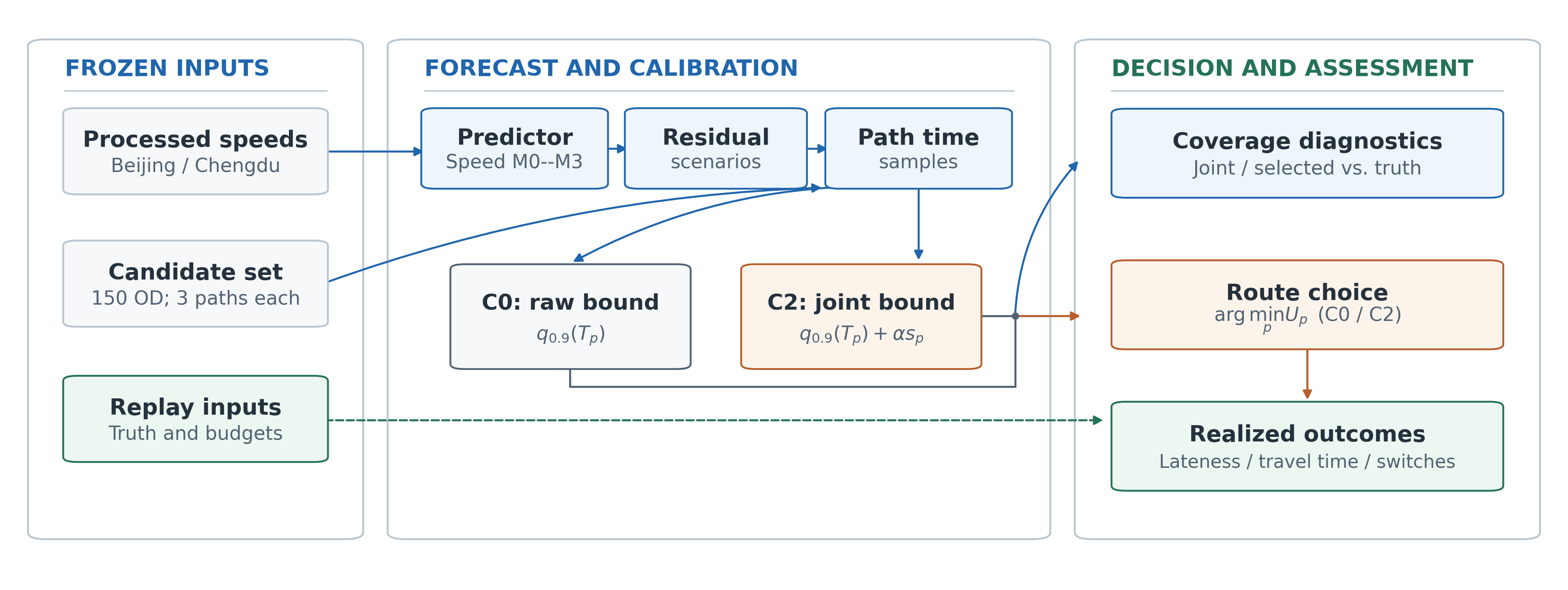}
  \caption{Framework of the frozen proxy-task evaluation. Processed speed data and fixed candidate paths feed the forecast and scenario stages. Path-time samples yield C0 raw bounds and C2 jointly calibrated bounds, whose coefficient is fitted on calibration days. Both bounds are assessed for joint and selected-path coverage against replay truth and passed through the specified minimum-upper-bound route interface. The dashed path carries replay truth and travel-time budgets to outcome evaluation; these are not available to the route selector. Coverage diagnostics and realized lateness, travel time, and switching are reported separately.}
  \label{fig:chain}
\end{figure}

Figure~\ref{fig:chain} summarizes the evaluation chain. The three-panel layout separates frozen inputs, forecast and calibration, and downstream assessment. Replay truth and budgets enter evaluation without entering route selection.

\subsection{Bounds and coverage targets}
The raw scenario quantile is C0:
\begin{equation}\label{eq:c0}
U^{(0)}_{o,p}=\qninety(T_{o,p}).
\end{equation}
C1 is a per-path calibration branch used in the broader experiment protocol. The primary comparison uses C2, a joint candidate-set upper-bound calibration:
\begin{equation}\label{eq:c2}
U^{(2)}_{o,p}=\max\{\qninety(T_{o,p})+\alpha s_{o,p},,0\}.
\end{equation}
The coefficient $\alpha$ is estimated only on the calibration period and fixed for testing. C2 targets simultaneous upper coverage of the three candidate paths for an origin--destination/departure observation; it is not a claim of independent 90\% coverage for every link or path.

Joint candidate-set coverage is the fraction of observations for which all three realized candidate-path times fall below their respective bounds. Selected-path coverage is the fraction for which the route chosen by the relevant upper-bound interface is covered. We report the absolute level and absolute deviation from the 90\% nominal target for joint coverage. A rise in selected-path coverage can represent greater conservatism, and need not mean closer alignment with 90\%.

\subsection{Decision interfaces and realized losses}
For fixed travel-time budget $B_o(t)$, C0 and C2 choose
\begin{equation}\label{eq:choice}
\hat p^{(j)}=\arg\min_{p\in\mathcal{P}_o} U^{(j)}_{o,p},\qquad j\in\{0,2\}.
\end{equation}
Other interfaces minimize scenario mean or empirical lateness probability, or use persistence and training-history baselines. The oracle chooses the fastest candidate under replayed truth and serves only as a post hoc within-set benchmark; it is not deployable.

The primary realized loss is
\begin{equation}\label{eq:late}
L_{o,t}=\mathbf{1}\{T_{o,\hat p}(t)>B_o(t)\}.
\end{equation}
We also report mean travel time, mean tardiness, within-candidate-set regret, and route-change frequency. Positive C2--C0 changes in lateness or travel time are adverse. Under the common replay truth, the difference in mean travel time equals the difference in mean regret observation by observation. We therefore treat regret as an interpretation of oracle headroom rather than independent corroboration of the travel-time effect.

\section{Data and Frozen Proxy Tasks}\label{sec:data}

\subsection{Data sources and scope}
The Beijing task uses Q-Traffic v2 \citep{liao2018qtraffic,guo2019urban}; the Chengdu task uses the public Figshare v4 road-speed data \citep{guo2019urban_dataset}. The publicly released Beijing network has 45,148 roads; the modeling cache/subnetwork has 2,693 roads, and the formal candidate paths use 2,301 roads. The corresponding Chengdu counts are 5,943, 2,776, and 2,523. Both tasks use 150 origin--destination pairs, three paths per pair, and 14 test dates. Across those dates, Beijing has 784 departure instances (56 per test day) and Chengdu has 350 (25 per test day); these counts produce 117,600 and 52,500 repeated OD--departure observations per configuration, respectively (Table~\ref{tab:scope}).

\begin{table}[t]
\caption{Public-data and frozen-task scope. The final column counts repeated origin--destination/departure observations per formal configuration, not independent traffic observations.}
\label{tab:scope}
\centering\small
\begin{tabular}{l l r r r r r r r}
\toprule
City & Data & Public roads & Cache roads & Path roads & OD & $K$ & Days & Repeated obs. \\
\midrule
Beijing & Q-Traffic v2 & 45,148 & 2,693 & 2,301 & 150 & 3 & 14 & 117,600 \\
Chengdu & Figshare v4 & 5,943 & 2,776 & 2,523 & 150 & 3 & 14 & 52,500 \\
\bottomrule
\end{tabular}
\end{table}

The data are publicly processed speeds. Upstream smoothing, missing-value completion, directional sharing, and turning restrictions cannot be fully reconstructed from the released products. Consequently, our outcomes describe offline proxy route decisions on processed data. They do not measure the benefit of an online route-guidance deployment.

\subsection{Split, candidate task, and budgets}
Model training, calibration, and testing are separated according to a frozen protocol. Candidate paths, budgets, test truth, and residual sources were fixed before formal evaluation. The Beijing primary comparison aggregates three formal $B=500$ scenario seeds from S6. Chengdu uses one frozen formal run with 15 residual vectors from three residual days in each time window. A separate Beijing E168-CDF branch reuses the seed-2026 predictor, fully enumerates 168 validation-B residual vectors, and recomputes empirical quantiles and the C2 coefficient. It is a post-result mechanism diagnostic, not a reconstruction or replacement of the formal three-seed result.

Each origin--destination/departure observation has a fixed budget $B_o(t)$. A route is late if its replayed travel time exceeds that budget. The ``per 10,000 proxy decisions'' expression below is a standardized effect size, not a conversion to daily citywide traffic losses.

\section{Methods and Statistical Scope}\label{sec:methods}

\subsection{Predictors and scenarios}
\begin{figure}[t]
  \centering
  \includegraphics[width=\linewidth]{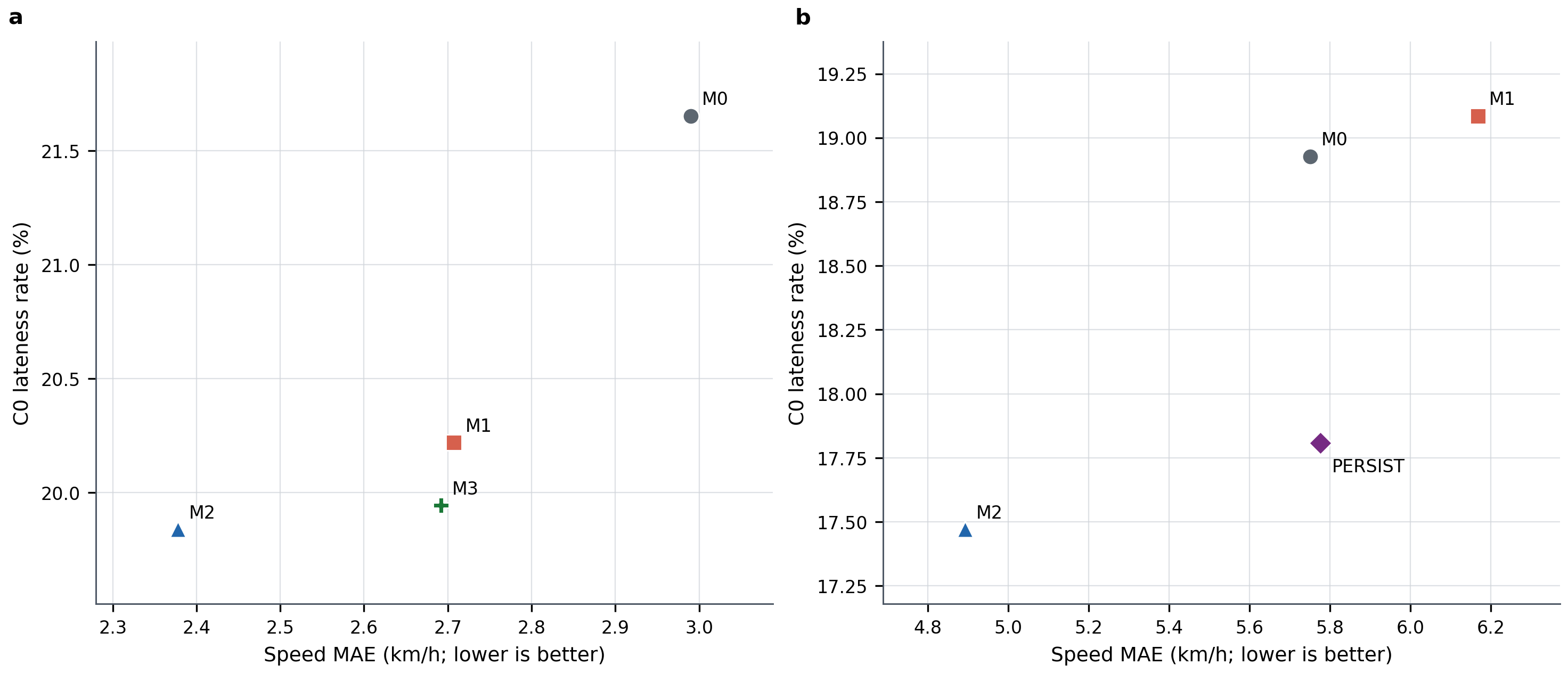}
  \caption{Within-city speed MAE and C0 lateness for frozen forecasting configurations: (a) Beijing and (b) Chengdu. Model replacement also changes residual and scenario distributions. The plotted relationship is descriptive and is not pooled across cities as a causal regression.}
  \label{fig:accuracy}
\end{figure}

The frozen study compares configurations M0--M3, without interpreting model labels as universal algorithm rankings. Each predictor feeds a link-speed-to-path-time scenario calculation, which in turn produces candidate-path quantiles and scales. Replacing a predictor can change point error, residual distributions, path scenarios, and route ordering simultaneously. Figure~\ref{fig:accuracy} is therefore a descriptive within-city comparison of speed MAE and C0 lateness; it is not a causal regression of lateness on MAE.

\subsection{Calibration and Monte Carlo support}
Calibration parameters use only the calibration period. The formal C2 protocol uses $B=500$ Monte Carlo scenarios in Beijing. In development, $B=200$ versus $B=500$ failed the prespecified 99\% route-agreement and 0.2-percentage-point lateness-tolerance checks, so we do not claim Monte Carlo convergence. A later B-grid diagnostic, conditional on a fixed library of 168 residual vectors, found mean C2--C0 lateness changes of 0.1843, 0.1620, 0.1528, 0.1493, and 0.1457 percentage points at $B=200,500,1000,2000,5000$, respectively. This diagnostic describes conditional resampling variation and does not replace the frozen formal $B=500$ protocol.

\subsection{Dependence and intervals}
For Beijing primary effects, we use a non-circular overlapping two-day moving-block bootstrap with 2,000 replicates and seed 20260920. Chengdu uses a circular two-day moving-block bootstrap with 2,000 replicates and seed 20260924. One-, two-, and three-day block lengths are indexed separately as sensitivity checks. Intervals describe variability within the frozen tasks and their 14 observed test dates; they do not give a finite-sample guarantee for an entire city. Residual-day deletion, complete 168-vector support, and numerical-path checks are supplemental diagnostics rather than independent external validation.

\section{Results}\label{sec:results}

\subsection{Improved point accuracy can accompany C0 decision gains}
For RQ1, Chengdu M2 reduces test speed MAE by 14.91\% relative to M0. With the same candidate set, budget, and C0 interface, lateness falls by 1.4571 percentage points and mean travel time falls by 0.1250 minutes; the two-day moving-block intervals exclude zero. Thus, point-forecast improvement can translate into decision improvement in this frozen Chengdu task. The comparison does not establish that a given reduction in MAE must cause a decision benefit: M2 also changes residual and path-scenario distributions, C0 minimizes only the raw $q_{0.9}$ score, and there is no user feedback.

\subsection{Coverage improvement and adverse effects under the specified interface}
\begin{figure}[t]
  \centering
  \includegraphics[width=\linewidth]{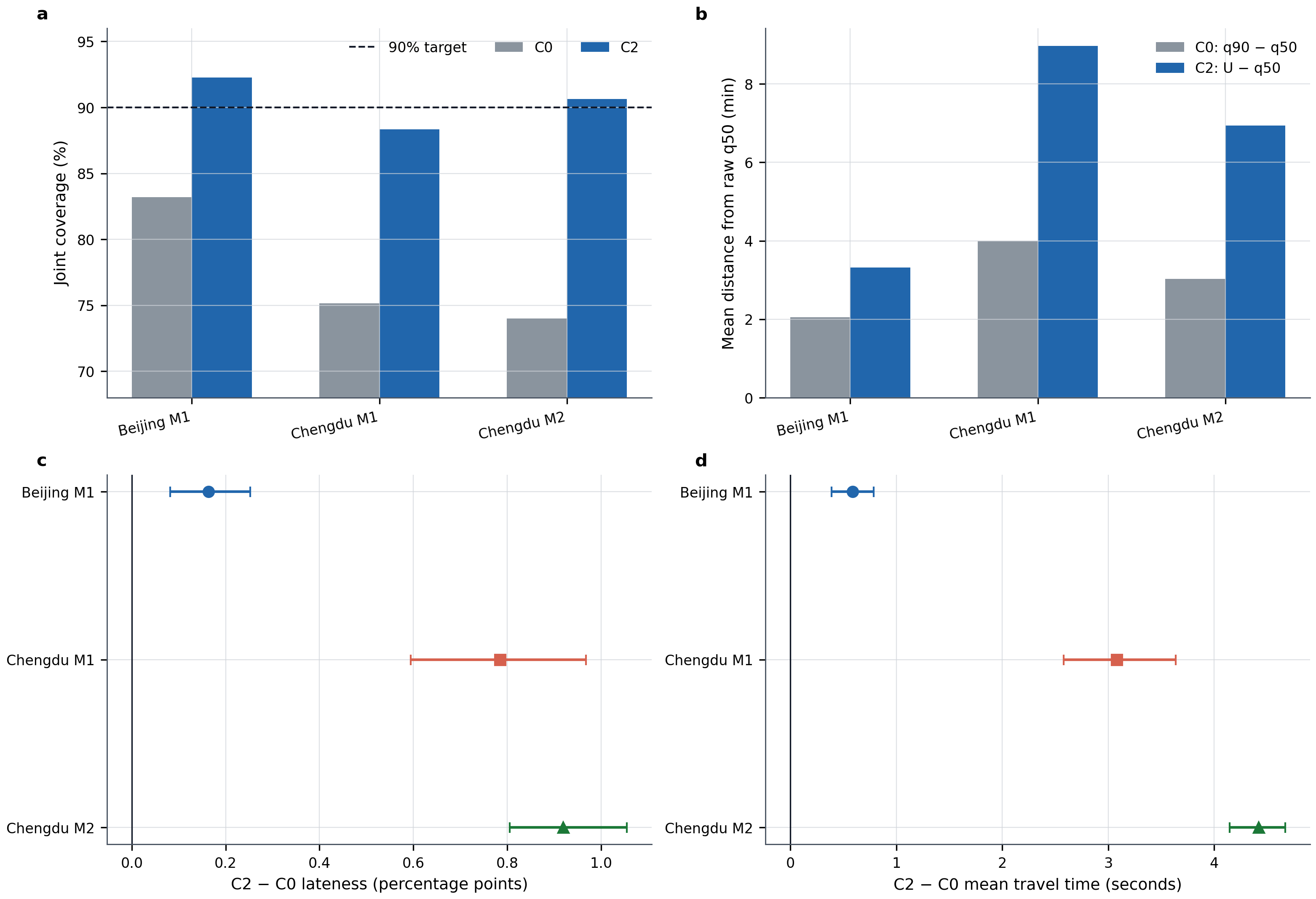}
  \caption{C2 versus C0 in the three formal comparisons. (a) Absolute joint candidate-set coverage against the 90\% target. (b) Mean distance of the raw $q_{0.9}$ and calibrated upper score from the raw median over the fixed three-path candidate sets. (c) Lateness changes and (d) mean travel-time changes with paired two-day moving-block intervals. Beijing uses non-circular blocks and Chengdu circular blocks. Positive decision changes are adverse.}
  \label{fig:tradeoff}
\end{figure}

Figure~\ref{fig:tradeoff} and Table~\ref{tab:tradeoff} summarize RQ2 and RQ3. C2 raises joint coverage in all three formal comparisons, while lateness and mean travel time increase under the minimum-upper-bound selector. The Beijing M1 joint coverage moves from 83.19\% to 92.26\%, reducing absolute deviation from 90\% from 6.81 to 2.26 percentage points. Chengdu M1 rises from 75.14\% to 88.33\%, reducing a coverage deficit but still falling short of 90\%. Chengdu M2 reaches 90.64\%, an absolute deviation of 0.64 percentage points. Selected-path coverage has a different interpretation: Beijing M1 rises from 87.71\% to 94.24\%, increasing its deviation from 90\% from 2.29 to 4.24 percentage points and indicating a more conservative selected-path bound.

\begin{table}[t]
\caption{Joint and selected-path coverage, fixed-candidate upper-score distance, and realized C2--C0 effects in the formal comparisons. Coverage and distance cells show C0 $\rightarrow$ C2. Intervals are paired two-day moving-block intervals.}
\label{tab:tradeoff}
\centering\footnotesize
\begin{tabular}{@{}l c c c c c@{}}
\toprule
Configuration & Joint cov. (\%) & Selected cov. (\%) & $U-\qfifty$ (min) & $\Delta$ late (pp) & $\Delta$ time (s) \\
\midrule
Beijing M1 & 83.19 $\to$ 92.26 & 87.71 $\to$ 94.24 & 2.049 $\to$ 3.324 & $+0.1633$ & $+0.588$ \\
Chengdu M1 & 75.14 $\to$ 88.33 & 79.50 $\to$ 90.25 & 3.999 $\to$ 8.971 & $+0.7848$ & $+3.082$ \\
Chengdu M2 & 74.01 $\to$ 90.64 & 81.12 $\to$ 92.99 & 3.025 $\to$ 6.942 & $+0.9200$ & $+4.418$ \\
\bottomrule
\end{tabular}
\vspace{2pt}
\parbox{\linewidth}{\scriptsize Lateness 95\% intervals (pp): Beijing M1 [0.0822, 0.2520], Chengdu M1 [0.5943, 0.9677], Chengdu M2 [0.8057, 1.0553]. Travel-time 95\% intervals (s): [0.390, 0.788], [2.578, 3.638], and [4.146, 4.672], respectively. Route-change rates are 2.89\%, 19.10\%, and 16.62\%.}
\end{table}

Raw $\qninety-\qfifty$ is not the calibrated expansion. Over fixed candidate sets, Beijing M1 has a raw distance of 2.049 minutes and a C2 $U-\qfifty$ distance of 3.324 minutes. The corresponding Chengdu M1 distances are 3.999 and 8.971 minutes; for Chengdu M2 they are 3.025 and 6.942 minutes. The additional upper-score penalty matters for traffic decisions only through its effect on candidate rankings. The observed lateness changes correspond to approximately 16.3, 78.5, and 92.0 additional late outcomes per 10,000 proxy decisions; we describe them as consistent but limited adverse changes within these tasks, not as severe system-wide efficiency losses.

\subsection{Decision interfaces produce different outcomes}
Table~\ref{tab:interfaces} compares alternative rules using aligned scenarios and budgets. In the Beijing E168-CDF supplemental branch, the lateness rates are 20.21\% for raw-$q_{0.9}$ selection, 20.35\% for C2 upper-score selection, 20.07\% for scenario-mean selection, and 20.04\% for empirical-lateness-probability selection. The nondeployable oracle reaches 19.36\%. C2 raises mean travel time relative to raw-$q_{0.9}$ selection by about 0.516 seconds in that supplemental branch.

In the formal Chengdu M2 run, the same respective lateness rates are 17.47\% for C0 and 18.39\% for C2; the scenario-mean, persistence/mean, historical-mean, and oracle baselines reach 16.81\%, 17.10\%, 17.42\%, and 12.83\%. The oracle reveals remaining headroom within the candidate set and must not be interpreted as a deployable strategy. These contrasts demonstrate that a coverage objective, a mean-time objective, and a lateness objective need not select the same path.

\begin{table}[t]
\caption{Decision interfaces, simple baselines, and candidate-set oracle headroom. Beijing results are from the post-result E168-CDF supplemental branch; Chengdu M2 results are formal. The oracle uses realized replay truth and is not deployable.}
\label{tab:interfaces}
\centering\footnotesize
\begin{tabular}{@{}l l r r r r@{}}
\toprule
Branch & Strategy & Late (\%) & Time (min) & Tardy (min) & Regret (min) \\
\midrule
Beijing E168 & Raw $q_{0.9}$ & 20.2134 & 11.5095 & 0.3187 & 0.0522 \\
 & C2 upper & 20.3452 & 11.5181 & 0.3224 & 0.0608 \\
 & Scenario mean & 20.0672 & 11.4997 & 0.3152 & 0.0424 \\
 & Empirical late prob. & 20.0383 & 11.5071 & 0.3164 & 0.0498 \\
 & Oracle & 19.3605 & 11.4573 & --- & 0 \\
\midrule
Chengdu M2 & Raw $q_{0.9}$ & 17.4705 & 18.4554 & 0.3385 & 0.4096 \\
 & C2 upper & 18.3905 & 18.5291 & 0.3537 & 0.4832 \\
 & Scenario mean & 16.8095 & 18.3911 & 0.3302 & 0.3453 \\
 & Persistence/mean & 17.1010 & 18.4121 & 0.3303 & 0.3663 \\
 & Historical mean & 17.4229 & 18.4742 & 0.3531 & 0.4283 \\
 & Oracle & 12.8305 & 18.0458 & 0.2498 & 0 \\
\bottomrule
\end{tabular}
\end{table}

\subsection{Route-switch and boundary decomposition}
\begin{figure}[t]
  \centering
  \includegraphics[width=\linewidth]{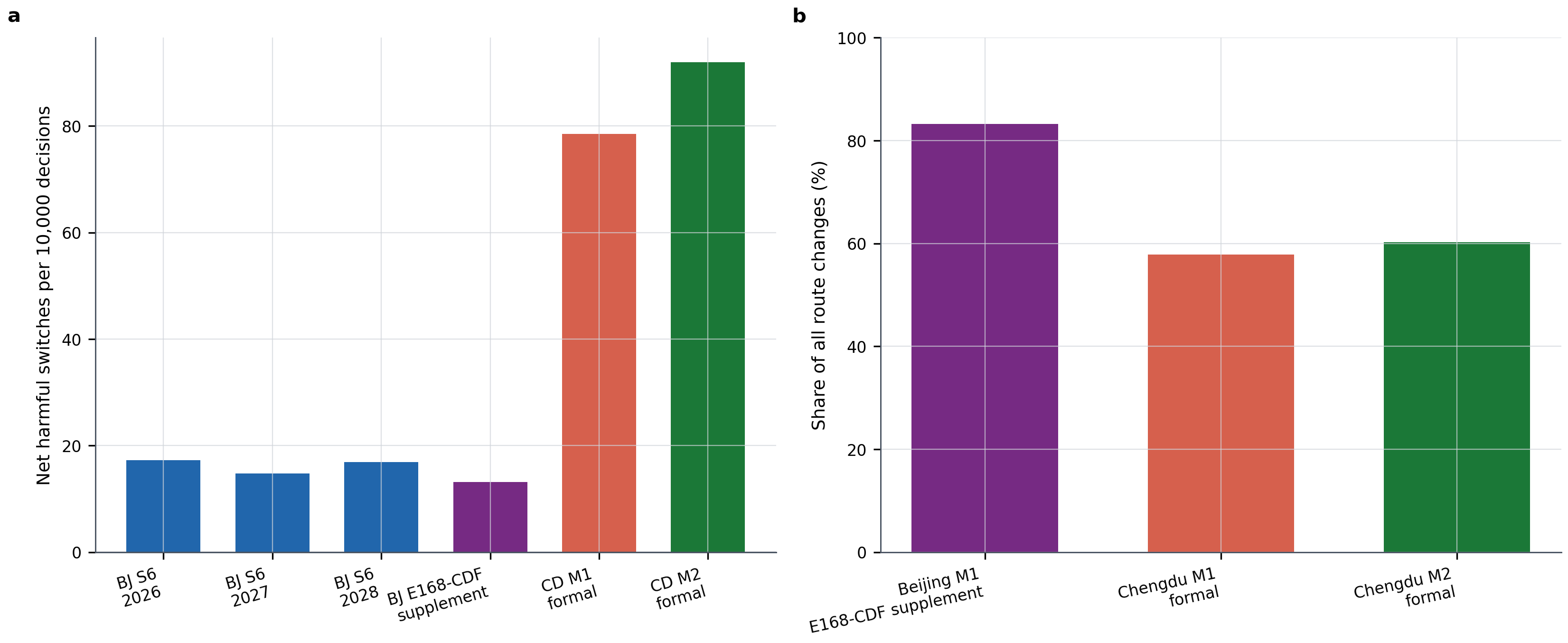}
  \caption{Formal and supplemental branch bridge and decision-boundary concentration. (a) Net harmful route switches per 10,000 proxy decisions for three formal Beijing S6 seeds, the separate Beijing E168-CDF supplement, and two formal Chengdu configurations. (b) Share of switches in the lowest quartile of the raw-$q_{0.9}$ gap; the Beijing value belongs only to E168-CDF.}
  \label{fig:switches}
\end{figure}

Figure~\ref{fig:switches} and Table~\ref{tab:switches} keep the formal Beijing branches separate from E168-CDF. The three formal S6 seeds yield 203, 174, and 199 net harmful switches, totaling 576 across 352,800 configuration observations, or 16.33 per 10,000. E168-CDF yields 155 net harmful switches, or 13.18 per 10,000, on its own 117,600 observations. The latter is not an exact reconstruction of formal seed 2026. Formal Chengdu M1 and M2 yield 412 and 483 net harmful switches.

\begin{table}[t]
\caption{Route-switch accounting by branch. Harmful means on-time under C0 and late under C2; beneficial means late under C0 and on-time under C2. The Beijing pooled row aggregates three formal configurations only.}
\label{tab:switches}
\centering\footnotesize
\begin{tabular}{@{}l r r r r r r@{}}
\toprule
Branch & Obs. & Switches & Harmful & Beneficial & Net harmful & Per 10,000 \\
\midrule
Beijing S6 seed 2026 & 117,600 & 3,372 & 235 & 32 & 203 & 17.26 \\
Beijing S6 seed 2027 & 117,600 & 3,389 & 207 & 33 & 174 & 14.80 \\
Beijing S6 seed 2028 & 117,600 & 3,418 & 236 & 37 & 199 & 16.92 \\
Beijing S6 pooled & 352,800 & 10,179 & 678 & 102 & 576 & 16.33 \\
Beijing E168-CDF & 117,600 & 3,197 & 193 & 38 & 155 & 13.18 \\
Chengdu M1 formal & 52,500 & 10,026 & 785 & 373 & 412 & 78.48 \\
Chengdu M2 formal & 52,500 & 8,728 & 756 & 273 & 483 & 92.00 \\
\bottomrule
\end{tabular}
\end{table}

The net counts equal harmful minus beneficial switches. Across the three Beijing formal seeds there are 678 harmful and 102 beneficial switches; E168-CDF has 193 and 38; Chengdu M1 has 785 and 373; Chengdu M2 has 756 and 273. This accounting explains the sign of the lateness difference within each frozen comparison. It is not an individual-level causal effect of route switching.

Switches concentrate close to the C0 decision boundary. The lowest quartile of the gap between the two smallest raw upper scores contains 83.2\% of Beijing E168-CDF switches, 57.9\% of formal Chengdu M1 switches, and 60.3\% of formal Chengdu M2 switches. Among harmful switches, approximately 58.3\%--60.6\% choose a longer path; hence path length provides only a partial explanation. Supplementary Figures~\ref{fig:switch-anatomy} and \ref{fig:heterogeneity} show additional branch-labeled switch and heterogeneity diagnostics.

\section{Robustness, Diagnostics, and Limitations}\label{sec:robustness}

\subsection{Completed checks}
Table~\ref{tab:robust} indexes completed sensitivity and stress tests. For Beijing M1, one-, two-, and three-day block intervals retain the C2--C0 point estimate of 0.1633 percentage points while changing interval width. Independent-link residual sampling preserves the adverse direction with a different magnitude. Replacing the path-dependent scale with a constant does not alter rankings and yields a zero decision effect. Quantile targets $q=0.80$, $0.90$, and $0.95$ produce lateness changes of 0.0604, 0.1633, and 0.3160 percentage points. Budget multipliers 0.9, 1.0, and 1.1 retain the adverse direction.

Candidate-set-size checks with $K=2,3,5$ also retain an adverse direction, but their $K=3$ and $K=5$ versions use 144 common feasible origin--destination pairs and an expanded road model, so they are not directly comparable with the frozen 150-pair primary task. Length, overlap, and time-window groupings were defined after protocol freeze and are exploratory. The 15-minute observation-delay check is a deployment-mismatch stress test; the released data cannot identify the causal effects of upstream completion or direction approximation. A day-maximum-score variant uses only 12 calibration-day maxima and is treated as a conservative stress test, not independent confirmation.

Full-support reconstruction on the fixed 168-vector Beijing library reproduces saved path statistics, calibration coefficients, and choices exactly. It removes resampling noise conditional on that support, but does not add traffic dates. In a residual-day deletion diagnostic retaining 112 vectors, lateness changes range from 0.1531 to 0.2185 percentage points. Numerical-path checks show a very small extended-lookup fraction, while speed-floor handling affects some scenarios. These checks bound numerical and residual-library concerns; they do not supply new external validation or causal evidence.

\subsection{Limits of interpretation}
First, upstream smoothing, completion, and directional sharing in the public speed products may alter the true speed process; original missingness masks and direction-specific information are unavailable. Second, only 14 test dates are observed and origin--destination/departure observations repeat within days. Moving-block intervals address part of the date dependence but cannot guarantee citywide generalization. Third, paths, budgets, and replay truth define a proxy task without user rerouting, congestion feedback, or behavioral response. Fourth, complete support in Beijing and the 15-vector within-window Chengdu support remove only conditional sampling variation, not questions about residual representativeness. Fifth, no randomized intervention identifies causal effects of the predictor, calibrator, or route change on individual lateness.

Accordingly, the supported claim is that joint upper-bound coverage and realized loss under a specific minimum-upper-bound interface separate in two frozen proxy tasks on processed public speed data. The result cannot be directly extrapolated to other cities, candidate-set sizes, calibration methods, real-time driving, or policy benefit.

\section{Discussion}\label{sec:discussion}

\subsection{Point accuracy and probabilistic coverage are distinct layers}
Two comparisons point in different directions. Chengdu M2 has lower point-prediction error than M0 and better decisions under C0. C2 raises joint coverage relative to C0, yet worsens lateness and mean travel time in the three formal comparisons. These results involve different interventions on the evaluation chain. Predictor replacement changes forecast centers and residual structure; C2 expands upper scores on existing scenarios and may reorder paths. The effects should be interpreted through their respective interfaces.

\subsection{Joint coverage is not a surrogate for decision benefit}
Joint coverage asks whether all candidate-path bounds cover replayed truth. Route choice asks whether the selected path is fast and meets its budget. If upper-score expansion differs among candidate paths, calibration can move a slightly faster but more uncertain path below a competitor in the score ranking. Because many changes occur when raw $q_{0.9}$ scores are nearly tied, modest ranking reversals can yield visible differences in realized losses. Joint coverage remains a useful reliability statistic, but its improvement is not a substitute for measuring the downstream objective.

\subsection{Implications for system evaluation}
A traffic prediction system should report at least three layers: forecast error and probabilistic scores; joint candidate-set coverage and bound expansion; and route-level lateness, travel time, budget exceedance, and switches. The route interface should be assessed against the operational loss of interest. A service concerned with lateness probability should evaluate the lateness-probability interface directly; one concerned with mean travel time cannot select a predictor solely on coverage. Our results do not establish a universally optimal interface. They show how layered evaluation reveals a mismatch between the reported probability target and the realized route outcome.

\section{Conclusion}\label{sec:conclusion}

We presented a layered framework for evaluating traffic uncertainty in route decisions and applied it to two frozen, public-data proxy tasks. In Chengdu, an M2--M0 reduction of 14.91\% in speed MAE accompanies lower lateness and mean travel time under the fixed C0 interface. Joint candidate-set calibration raises coverage in Beijing M1, Chengdu M1, and Chengdu M2, but all three formal C2--C0 comparisons have positive, limited changes in lateness and travel time under the specified minimum-upper-bound rule. Route changes cluster near raw-$q_{0.9}$ decision boundaries, and harmful changes outnumber beneficial ones. This decomposition explains the frozen-task outcomes without identifying a behavioral or network-level causal mechanism. Future work should examine the evaluation chain with live data, direction and turn constraints, user response, and additional cities.

\section*{Reproducibility Statement}
The core evidence repository is available at \url{https://github.com/ljnvs/traffic_2.git}. It contains the frozen aggregate result tables, day-level outcomes, bootstrap indices, and a script that verifies the headline contrasts and intervals. The underlying processed Beijing and Chengdu road-speed data remain available from their original providers \citep{liao2018qtraffic,guo2019urban_dataset}. The repository supports result-level checks; it does not redistribute the raw speed archives or provide an end-to-end rerun of model training and proxy-task construction.

\section*{Ethics and Competing Interests}
The study uses publicly released, de-identified road-speed data and does not analyze individual trajectories or human participants. The authors declare no competing interests related to this work.

\clearpage
\appendix
\renewcommand{\thefigure}{S\arabic{figure}}
\renewcommand{\thetable}{S\arabic{table}}
\renewcommand{\theHfigure}{S\arabic{figure}}
\renewcommand{\theHtable}{S\arabic{table}}
\setcounter{figure}{0}
\setcounter{table}{0}
\section{Supplementary Figures and Evaluation Definitions}\label{app:supp}

The supplementary figures and tables below retain the formal and post-result branch labels used in the frozen analysis.

\begin{figure}[H]
  \centering
  \includegraphics[width=\linewidth]{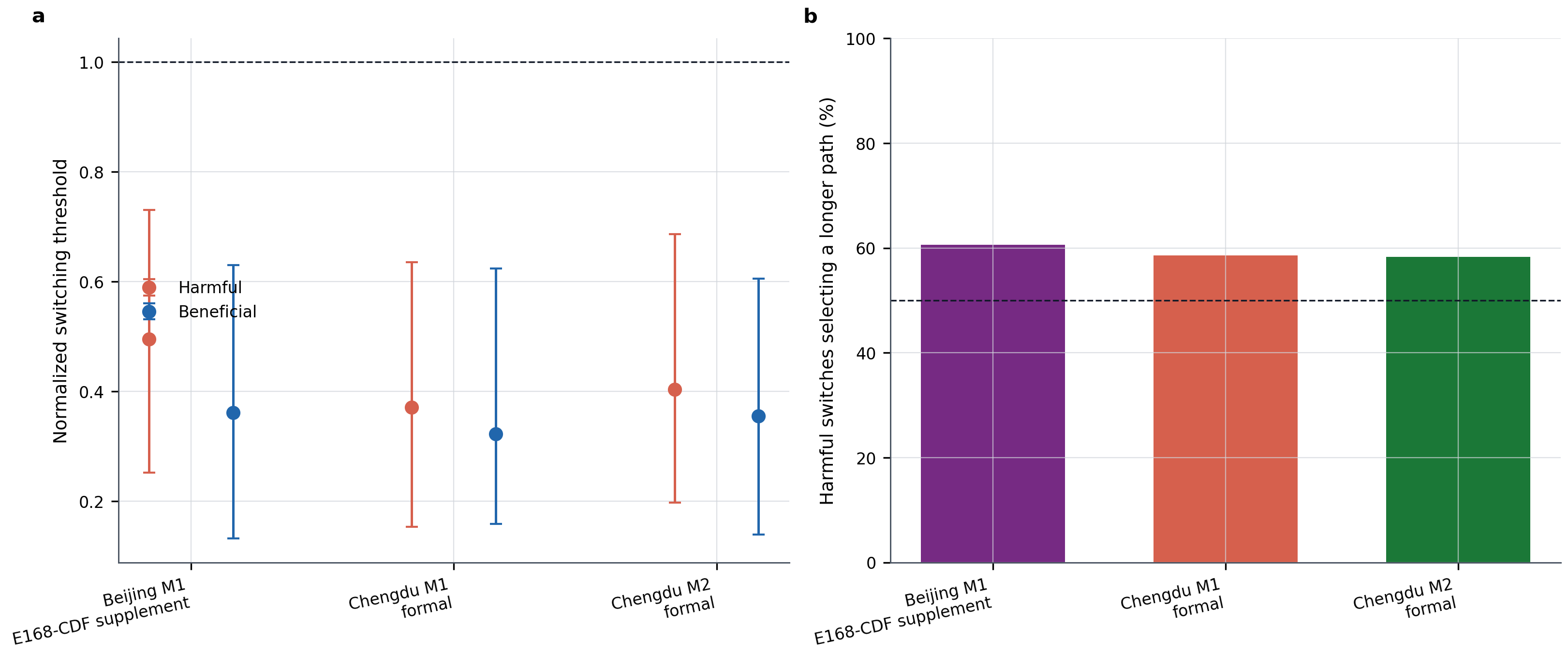}
  \caption{Branch-labeled switch anatomy. (a) Median and interquartile range of the normalized switching threshold for harmful and beneficial switches. (b) Share of harmful switches that choose a longer path. Beijing belongs to the E168-CDF supplement and Chengdu to formal comparisons. These are descriptive outcomes within the proxy tasks.}
  \label{fig:switch-anatomy}
\end{figure}

\begin{figure}[H]
  \centering
  \includegraphics[width=\linewidth]{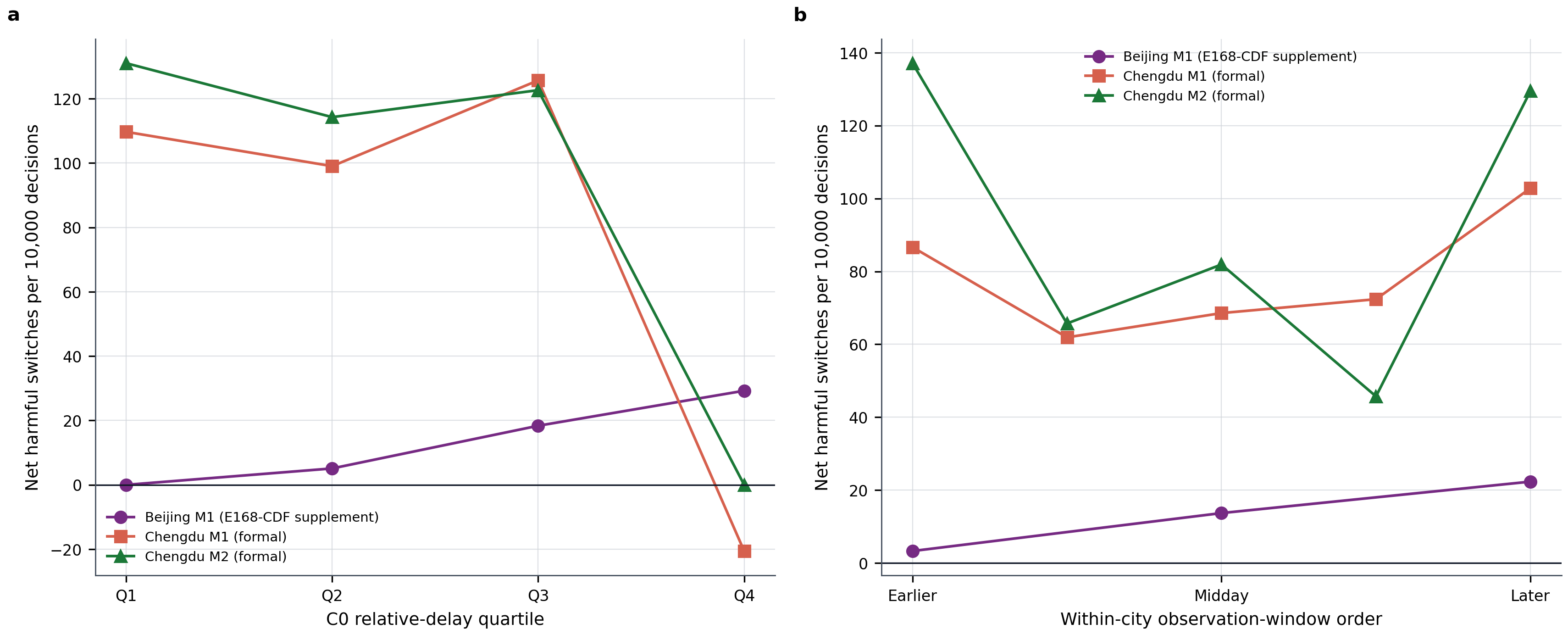}
  \caption{Branch-labeled heterogeneity of net harmful switches per 10,000 decisions by (a) C0 relative-delay quartile and (b) within-city observation-window order. Patterns are descriptive and are not uniform across cities; they do not establish time-of-day intervention rules.}
  \label{fig:heterogeneity}
\end{figure}

\begin{table}[H]
\caption{Definitions and interpretation of the primary evaluation quantities.}
\label{tab:definitions}
\centering\footnotesize
\begin{tabularx}{\linewidth}{@{}l X X@{}}
\toprule
Quantity & Definition & Interpretation \\
\midrule
Speed MAE & Mean absolute road-speed forecast error on test records & Point-forecast accuracy; lower is better. \\
Joint coverage & Fraction of OD--departures for which all three candidate travel times do not exceed their bounds & Compare absolute level and distance from 90\%. \\
Selected coverage & Coverage of the path chosen under the respective upper-score interface & Higher values can reflect conservatism. \\
Upper-score distance & Candidate-set mean $U-\qfifty$, using a fixed three-path set & Reports score expansion, not only raw $q_{0.9}-q_{0.5}$. \\
Lateness & Fraction with replayed selected-path time above fixed budget & Positive C2--C0 change is adverse. \\
Mean travel time & Mean replayed time of selected paths & Its C2--C0 difference equals regret difference under common truth. \\
Net harmful switches & On-time-to-late switches minus late-to-on-time switches & Accounting of the lateness-rate difference, not a causal mechanism. \\
\bottomrule
\end{tabularx}
\end{table}

\begin{table}[H]
\caption{Statistical unit and interval scope. Repeated route decisions within the same date are not independent traffic days.}
\label{tab:statistics}
\centering\footnotesize
\begin{tabularx}{\linewidth}{@{}l X X@{}}
\toprule
Task & Bootstrap construction & Scope \\
\midrule
Beijing formal & Non-circular, overlapping two-day moving blocks; 2,000 replicates; seed 20260920 & Paired effects over 14 observed test dates and three formal B=500 seeds. \\
Chengdu formal & Circular two-day moving blocks; 2,000 replicates; seed 20260924 & Paired effects over 14 observed test dates in one frozen run. \\
Sensitivity & One- and three-day block lengths examined separately & Interval-width check, not city-population inference. \\
\bottomrule
\end{tabularx}
\end{table}

\begin{table}[H]
\caption{Selected completed robustness and diagnostic checks. The listed checks retain their original formal or supplemental status.}
\label{tab:robust}
\centering\scriptsize
\begin{tabularx}{\linewidth}{@{}l X X@{}}
\toprule
Check & Variation & Main finding or interpretation \\
\midrule
Block length & One-, two-, three-day blocks & Beijing point effect remains +0.1633 pp; interval width varies. \\
Residual dependence & Independent-link sampling & Adverse direction remains, with changed magnitude. \\
Constant scale & Replace path-dependent $s_{o,p}$ & No route reordering; zero decision effect. \\
Quantile target & $q=0.80,0.90,0.95$ & Lateness changes +0.0604, +0.1633, +0.3160 pp. \\
Budget & Multipliers 0.9, 1.0, 1.1 & Adverse direction remains. \\
Candidate set & $K=2,3,5$ & Adverse direction; expanded task uses 144 common feasible OD pairs and is not directly comparable with the primary task. \\
E168 full support & Enumerate 168 residual vectors & Exact conditional reconstruction; no additional test dates. \\
B grid & $B=200$ to 5000 conditional on fixed residual library & Mean adverse effect declines from 0.1843 to 0.1457 pp; formal $B=500$ retained. \\
Residual-day deletion & Retain 112 vectors & Adverse lateness change ranges 0.1531--0.2185 pp. \\
Other pressure checks & Observation delay, day-max score, numerical path & Identify deployment and computation boundaries; not independent external validation. \\
\bottomrule
\end{tabularx}
\end{table}

\clearpage
\small
\setlength{\bibsep}{3pt}
\bibliographystyle{unsrtnat}
\bibliography{references_research}

@inproceedings{liao2018qtraffic,
  author    = {Binbing Liao and Jingqing Zhang and Chao Wu and Douglas McIlwraith and Tong Chen and Shengwen Yang and Yike Guo and Fei Wu},
  title     = {Deep Sequence Learning with Auxiliary Information for Traffic Prediction},
  booktitle = {Proceedings of the 24th ACM SIGKDD International Conference on Knowledge Discovery and Data Mining},
  pages     = {537--546},
  year      = {2018},
  doi       = {10.1145/3219819.3219895}
}

@article{guo2019urban,
  author  = {Feng Guo and Dongqing Zhang and Yucheng Dong and Zhaoxia Guo},
  title   = {Urban Link Travel Speed Dataset from a Megacity Road Network},
  journal = {Scientific Data},
  volume  = {6},
  number  = {1},
  pages   = {61},
  year    = {2019},
  doi     = {10.1038/s41597-019-0060-3}
}

@misc{guo2019urban_dataset,
  author    = {Feng Guo and Dongqing Zhang and Yucheng Dong and Zhaoxia Guo},
  title     = {Urban Link Travel Speed Dataset from a Megacity Road Network},
  publisher = {figshare},
  year      = {2019},
  version   = {4},
  doi       = {10.6084/m9.figshare.7140209.v4},
  note      = {CC0 1.0}
}

@article{qian2024unified,
  author  = {Weizhu Qian and Yan Zhao and Dalin Zhang and Bowei Chen and Kai Zheng and Xiaofang Zhou},
  title   = {Towards a Unified Understanding of Uncertainty Quantification in Traffic Flow Forecasting},
  journal = {IEEE Transactions on Knowledge and Data Engineering},
  volume  = {36},
  number  = {5},
  pages   = {2239--2256},
  year    = {2024},
  doi     = {10.1109/TKDE.2023.3312261}
}

@article{wu2024adaptive,
  author  = {Ying Wu and Yongchao Ye and Adnan Zeb and James Jianqiao Yu and Zheng Wang},
  title   = {Adaptive Modeling of Uncertainties for Traffic Forecasting},
  journal = {IEEE Transactions on Intelligent Transportation Systems},
  volume  = {25},
  number  = {5},
  pages   = {4427--4442},
  year    = {2024},
  doi     = {10.1109/TITS.2023.3327100}
}

@misc{xiong2026stochasticity,
  author        = {Weijiang Xiong and Robert Fonod and Nikolas Geroliminis},
  title         = {Unveiling Stochasticity: Universal Multi-modal Probabilistic Modeling for Traffic Forecasting},
  year          = {2026},
  eprint        = {2604.16084},
  archiveprefix = {arXiv},
  primaryclass  = {cs.LG},
  note          = {arXiv v2, updated 2026-07-03; no journal reference or DOI recorded on 2026-09-21}
}

@inproceedings{romano2019conformalized,
  author    = {Yaniv Romano and Evan Patterson and Emmanuel Candès},
  title     = {Conformalized Quantile Regression},
  booktitle = {Advances in Neural Information Processing Systems},
  volume    = {32},
  year      = {2019},
  url       = {https://proceedings.neurips.cc/paper/2019/hash/5103c3584b063c431bd1268e9b5e76fb-Abstract.html}
}

@inproceedings{stankeviciute2021conformal,
  author    = {Kamile Stankeviciute and Ahmed M. Alaa and Mihaela van der Schaar},
  title     = {Conformal Time-series Forecasting},
  booktitle = {Advances in Neural Information Processing Systems},
  volume    = {34},
  pages     = {6216--6228},
  year      = {2021},
  url       = {https://proceedings.neurips.cc/paper/2021/hash/312f1ba2a72318edaaa995a67835fad5-Abstract.html}
}

@inproceedings{gibbs2021adaptive,
  author    = {Isaac Gibbs and Emmanuel Candès},
  title     = {Adaptive Conformal Inference Under Distribution Shift},
  booktitle = {Advances in Neural Information Processing Systems},
  volume    = {34},
  pages     = {1660--1672},
  year      = {2021},
  url       = {https://proceedings.neurips.cc/paper/2021/hash/0d441de75945e5acbc865406fc9a2559-Abstract.html}
}

@inproceedings{xu2023sequential,
  author    = {Chen Xu and Yao Xie},
  title     = {Sequential Predictive Conformal Inference for Time Series},
  booktitle = {Proceedings of the 40th International Conference on Machine Learning},
  series    = {Proceedings of Machine Learning Research},
  volume    = {202},
  pages     = {38707--38727},
  year      = {2023},
  url       = {https://proceedings.mlr.press/v202/xu23r.html}
}

@inproceedings{cortesgomez2025utility,
  author    = {Santiago Cortes-Gomez and Carlos Patiño and Yewon Byun and Steven Wu and Eric Horvitz and Bryan Wilder},
  title     = {Utility-Directed Conformal Prediction: A Decision-Aware Framework for Actionable Uncertainty Quantification},
  booktitle = {International Conference on Learning Representations},
  pages     = {4227--4246},
  year      = {2025},
  url       = {https://proceedings.iclr.cc/paper_files/paper/2025/hash/0c6b452f1bbfb6905f6bac957d73b321-Abstract-Conference.html}
}

@inproceedings{kiyani2025decision,
  author    = {Shayan Kiyani and George J. Pappas and Aaron Roth and Hamed Hassani},
  title     = {Decision Theoretic Foundations for Conformal Prediction: Optimal Uncertainty Quantification for Risk-Averse Agents},
  booktitle = {Proceedings of the 42nd International Conference on Machine Learning},
  series    = {Proceedings of Machine Learning Research},
  volume    = {267},
  pages     = {30943--30965},
  year      = {2025},
  url       = {https://proceedings.mlr.press/v267/kiyani25a.html}
}

@inproceedings{donti2017task,
  author    = {Priya Donti and Brandon Amos and J. Zico Kolter},
  title     = {Task-based End-to-end Model Learning in Stochastic Optimization},
  booktitle = {Advances in Neural Information Processing Systems},
  volume    = {30},
  year      = {2017},
  url       = {https://papers.nips.cc/paper/2017/hash/3fc2c60b5782f641f76bcefc39fb2392-Abstract.html}
}

@inproceedings{wilder2019melding,
  author    = {Bryan Wilder and Bistra Dilkina and Milind Tambe},
  title     = {Melding the Data-Decisions Pipeline: Decision-Focused Learning for Combinatorial Optimization},
  booktitle = {Proceedings of the AAAI Conference on Artificial Intelligence},
  volume    = {33},
  pages     = {1658--1665},
  year      = {2019},
  doi       = {10.1609/aaai.v33i01.33011658}
}

@article{elmachtoub2022smart,
  author  = {Adam N. Elmachtoub and Paul Grigas},
  title   = {Smart ``Predict, then Optimize''},
  journal = {Management Science},
  volume  = {68},
  number  = {1},
  pages   = {9--26},
  year    = {2022},
  doi     = {10.1287/mnsc.2020.3922}
}

@article{samaranayake2012tractable,
  author  = {S. Samaranayake and S. Blandin and A. Bayen},
  title   = {A Tractable Class of Algorithms for Reliable Routing in Stochastic Networks},
  journal = {Transportation Research Part C: Emerging Technologies},
  volume  = {20},
  number  = {1},
  pages   = {199--217},
  year    = {2012},
  doi     = {10.1016/j.trc.2011.05.009}
}

@article{nikolova2014meanrisk,
  author  = {E. Nikolova and N. E. Stier-Moses},
  title   = {A Mean-Risk Model for the Traffic Assignment Problem with Stochastic Travel Times},
  journal = {Operations Research},
  volume  = {62},
  number  = {2},
  pages   = {366--382},
  year    = {2014},
  doi     = {10.1287/opre.2013.1246}
}

@article{huang2018trajectory,
  author  = {He Huang and Song Gao},
  title   = {Trajectory-Adaptive Routing in Dynamic Networks with Dependent Random Link Travel Times},
  journal = {Transportation Science},
  volume  = {52},
  number  = {1},
  pages   = {102--117},
  year    = {2018},
  doi     = {10.1287/trsc.2016.0691}
}

@article{filipovska2021reliable,
  author  = {Monika Filipovska and Hani S. Mahmassani},
  title   = {Reliable Trajectory-Adaptive Routing Strategies in Stochastic, Time-Varying Networks with Generalized Correlations},
  journal = {Transportation Research Part C: Emerging Technologies},
  volume  = {133},
  pages   = {103436},
  year    = {2021},
  doi     = {10.1016/j.trc.2021.103436}
}

@article{filipovska2021estimation,
  author  = {Monika Filipovska and Hani S. Mahmassani and Archak Mittal},
  title   = {Estimation of Path Travel Time Distributions in Stochastic Time-Varying Networks with Correlations},
  journal = {Transportation Research Record: Journal of the Transportation Research Board},
  volume  = {2675},
  number  = {11},
  pages   = {498--508},
  year    = {2021},
  doi     = {10.1177/03611981211018464}
}

\end{document}